\pdfoutput=1

\documentclass[11pt]{article}

\usepackage[final]{acl}

\usepackage{times}
\usepackage{latexsym}

\usepackage[T1]{fontenc}

\usepackage[utf8]{inputenc}

\usepackage{microtype}

\usepackage{inconsolata}

\usepackage{graphicx}
\usepackage{amsmath}
\usepackage{algorithm}
\usepackage{algorithmic}
\usepackage{latexsym}

\usepackage{microtype}

\usepackage{booktabs}
\usepackage{multirow}
\usepackage{mathtools}
\usepackage{subfigure}
\usepackage{balance}
\usepackage{color}
\usepackage{xcolor}
\usepackage{enumitem}
\usepackage{xspace}
\usepackage{setspace}
\usepackage{amssymb}
\usepackage{longtable}

\usepackage[show]{chato-notes}

\newcommand{\stitle}[1]{\vspace{1mm} \noindent {\bf #1}}

\newcommand{\model}{DLISC}

\title{Effective and Efficient Schema-aware Information Extraction Using On-Device Large Language Models}

\author{%
  Zhihao Wen, Sheng Liang, Yaxiong Wu, Yongyue Zhang, Yong Liu \\
  Huawei Noah’s Ark Lab\\
  \texttt{liangsheng16@huawei.com} \\
}

\begin{document}
\maketitle
\begin{abstract}

Information extraction (IE) plays a crucial role in natural language processing (NLP) by converting unstructured text into structured knowledge. 
Deploying computationally intensive large language models (LLMs) on resource-constrained devices for information extraction is challenging, particularly due to issues like hallucinations, limited context length, and high latency—especially when handling diverse extraction schemas.
To address these challenges, we propose a two-stage information extraction approach adapted for on-device LLMs, called \underline{D}ual-\underline{L}oRA with \underline{I}ncremental \underline{S}chema \underline{C}aching (DLISC), which enhances both schema identification and schema-aware extraction in terms of effectiveness and efficiency.
In particular, DLISC adopts an Identification LoRA module for retrieving the most relevant schemas to a given query, and an Extraction LoRA module for performing information extraction based on the previously selected schemas.
To accelerate extraction inference, Incremental Schema Caching is incorporated to reduce redundant computation, substantially improving efficiency.
Extensive experiments across multiple information extraction datasets demonstrate notable improvements in both effectiveness and efficiency.

\end{abstract}

\section{Introduction}

Information extraction (IE) is a core task in natural language processing (NLP) that aims to extract structured knowledge—such as entities, relations, and events—from unstructured text~\cite{xu2023large,deng2024information,yang2022survey}. 
Large language models (LLMs), with their powerful generalization abilities, have shown considerable promise in improving IE tasks~\cite{xu2023large,deng2024information}.
However, deploying LLMs on resource-constrained edge devices for information extraction is more challenging~\cite{xu2024device}, including hallucinations, limitations in context length, and high latency.
In particular, on-device LLMs face hallucinations due to insufficient task-specific tuning, while the need to include all extraction schemas in broad scenarios results in long inputs and high latency.

\begin{figure}[!t]
    \centering
    \includegraphics[width=0.99\linewidth]{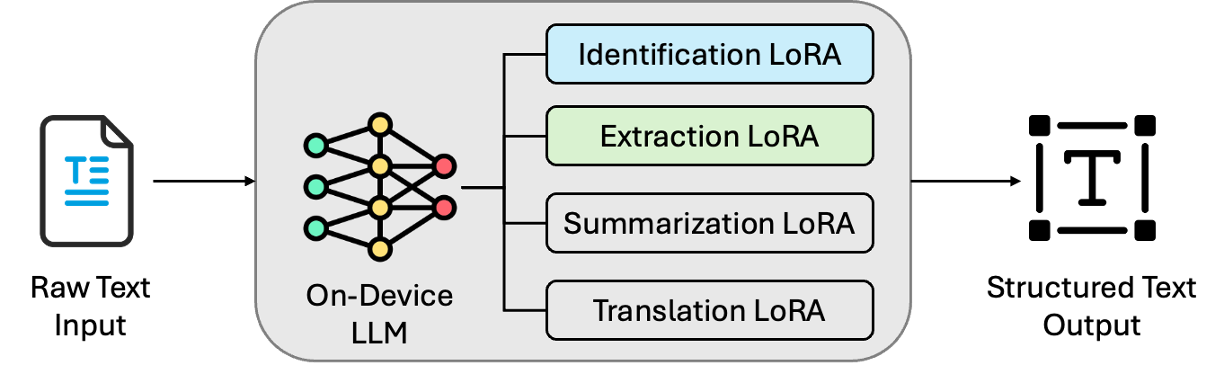}
    \caption{
    The LLM-Adapters architecture for deploying LLMs on edge devices with a single on-device LLM and multiple plug-in LoRA modules.}
    \label{fig:ondevice-llm}
    \vspace{-1\baselineskip}
\end{figure}

To mitigate these challenges, retrieval-augmented generation (RAG) methods have emerged as a promising solution, enhancing extraction accuracy by incorporating external knowledge~\cite{gao2023retrieval,Li2024FromMT}. 
For instance, \citet{shiri-2024-decompose} decompose event extraction into two subtasks: Event Detection (ED), which retrieves relevant event examples, and Event Argument Extraction (EAE), which extracts events based on the retrieved examples.
In addition, \citet{liang2025adaptive} propose Adaptive Schema-Aware Event Extraction (ASEE), a two-stage paradigm that decomposes the extraction task into schema matching and schema-augmented extraction.
ASEE leverages an extensive library of event extraction schemas, adaptively retrieving relevant schemas and assembling extraction prompts to improve accuracy and scalability.

\begin{figure*}[!ht]
    \centering
    \includegraphics[width=0.99\linewidth]{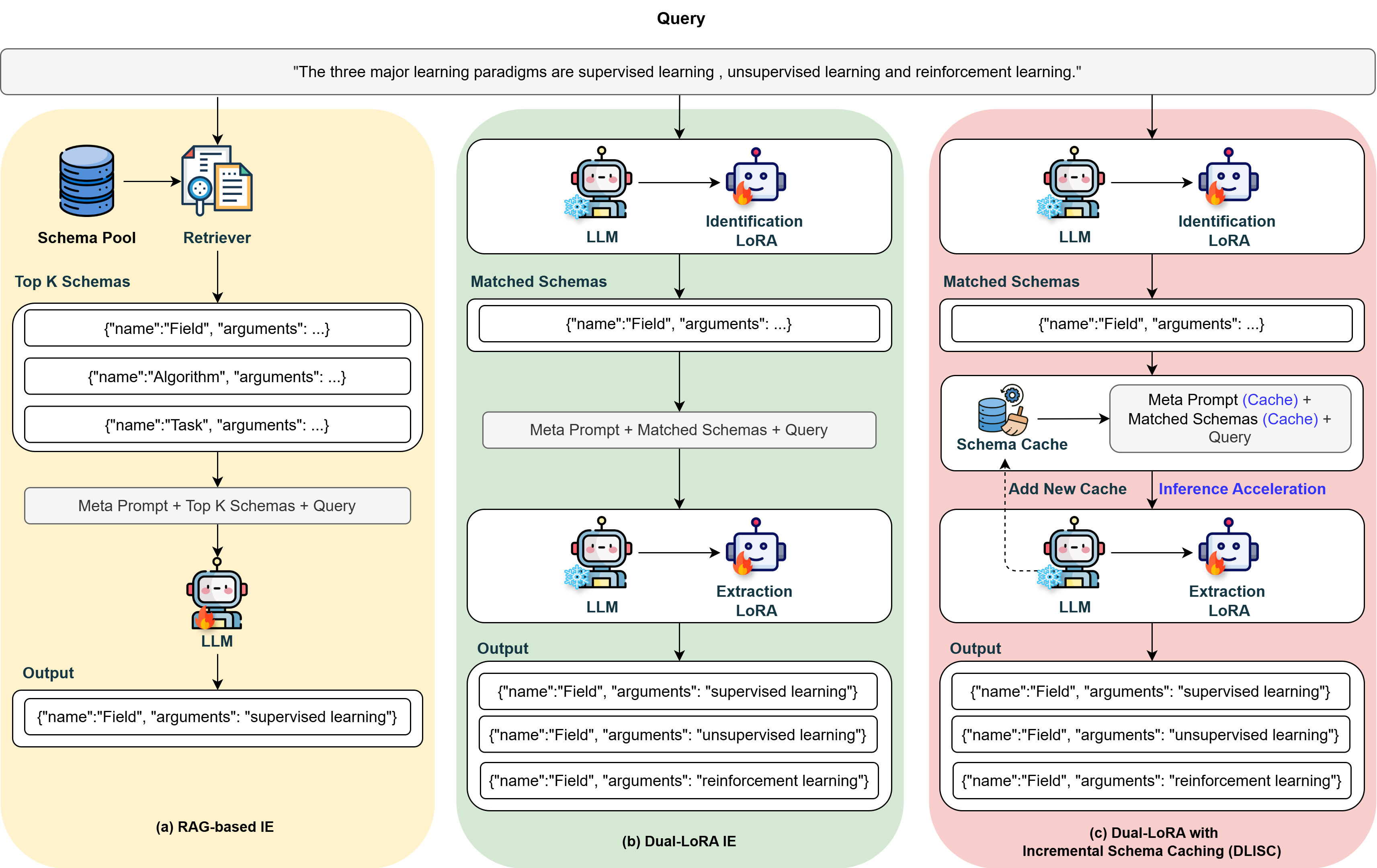}
    \caption{
    An illustrative comparison of (a) RAG-based IE with schema retrieval and top-K schema-aware extraction;
    (b) Dual-LoRA IE paradigm with schema identification and schema-aware extraction; 
    (c) Dual-LoRA with Incremental Schema Caching (DLISC) for further enhancing inference efficiency.}
    \label{fig:architecture}
    \vspace{-1\baselineskip}
\end{figure*}

Despite the progress of RAG-based methods in information extraction, fully leveraging the unique advantages of on-device LLMs~\cite{xu2024device,mehta2024openelm}—such as the LLM-Adapters architecture~\cite{hu-etal-2023-llm}—while enhancing extraction effectiveness and efficiency remains underexplored.
Figure~\ref{fig:ondevice-llm} illustrates the LLM-Adapters architecture for deploying large language models on edge devices. 
In this design, a single on-device LLM remains persistently loaded to maintain responsiveness, while multiple plug-in LoRA modules are selectively activated to support various task-specific adaptations with minimal resource overhead.

In this paper, we propose a two-stage information extraction approach adapted for on-device LLMs, called \underline{D}ual-\underline{L}oRA with \underline{I}ncremental \underline{S}chema \underline{C}aching (DLISC), which enhances both schema identification and schema-aware extraction in terms of effectiveness and efficiency.
In particular, DLISC adopts an Identification LoRA module for retrieving the most relevant schemas to a given query, and an Extraction LoRA module for performing information extraction based on the previously selected schemas.
To accelerate extraction inference, Incremental Schema Caching is incorporated to reduce redundant computation, notably improving efficiency.

\section{Methodology}
\label{sec:methodology}

To enhance the effectiveness and efficiency of on-device information extraction, we propose a two-stage information extraction approach adapted for on-device LLMs, called \underline{D}ual-\underline{L}oRA with \underline{I}ncremental \underline{S}chema \underline{C}aching (DLISC).
Figure~\ref{fig:architecture} presents the architecture of DLISC with an Identification LoRA for identifying the most relevant schemas to the query, an Extraction LoRA for performing the information extraction with the matched schemas, and Incremental Schema Caching for accelerating the extraction inference.

\stitle{Dual-LoRA Architecture.}
The Dual-LoRA information extraction architecture (as shown in Figure~\ref{fig:architecture} (b)) follows the RAG-based two-stage paradigm~\cite{liang2025adaptive} with an Identification LoRA ($\boldsymbol{\theta}_I$) and an Extraction LoRA ($\boldsymbol{\theta}_E$) based on the same LLM ($\boldsymbol{\theta}$).
During inference, these two parameter sets are ``merged'' with the LLM $(\boldsymbol{\theta})$, producing two distinct LLMs $(\boldsymbol{\theta}_I^\prime, \boldsymbol{\theta}_E^\prime)$, 
\begin{align}
    \boldsymbol{\theta}_I^\prime, \boldsymbol{\theta}_E^\prime = Merge(\boldsymbol{\theta}_I, \boldsymbol{\theta}), Merge(\boldsymbol{\theta}_E, \boldsymbol{\theta})
\end{align}
each serving identification and extraction functions. In addition, the Identification LoRA ($\boldsymbol{\theta}_I$) and Extraction LoRA ($\boldsymbol{\theta}_E$) can be optimized for improving the identification and extraction accuracy, respectively.

The raw text data, Query $Q$, is input into $\boldsymbol{\theta}_I^\prime$ with Identification Meta Prompt $M_{I}$ to identify the most relevant schemas, i.e. Matched Schemas $S$,
\begin{align}
    S = \boldsymbol{\theta}_I^\prime(M_{I}+Q).
\end{align}
The Matched Schemas $S$ are then concatenated with Extraction Meta Prompt $M_{E}$ and Query $Q$ as the prompt of $\boldsymbol{\theta}_E^\prime$,
\begin{align}
    R= & \boldsymbol{\theta}_E^\prime(M_{E}+S+Q),
\end{align}
where the returned structured results $R$ are the extracted information from the raw text data $Q$.

\stitle{Incremental Schema Caching Acceleration.}
To accelerating the extraction inference, we introduce Incremental Schema Caching (ISC) to the extraction process (as shown in Figure~\ref{fig:architecture} (c)), inspired by the Key-Value (KV) Cache mechanism~\cite{Luohe2024KeepTC} and Prompt Cache~\cite{gim2024prompt},
\begin{align}
    R= & \boldsymbol{\theta}_E^\prime(M_{E(Cache)}+S_{(Cache)}+Q),
\end{align}
where $M_{E(Cache)}$ and $S_{(Cache)}$ are the cached Extraction Meta Prompt and the cached Matched Schemas.
Specifically, we store the Extraction Meta Prompt $M_{E}$ in the cache when it first appears, and the Incremental Schema Caching (ISC) mechanism for the Matched Schemas works as follows:
\begin{itemize}[leftmargin=*, itemsep=-2pt, topsep=0pt, parsep=1pt] 
    \item When a Matched Schema $S$ is identified,  the schema is checked if it is already in the schema cache pool.
    \item If the schema cache \textbf{is} found, the cached schema is directly returned and used for accelerating the extraction inference.
    \item If the schema cache \textbf{is not} found, the Matched Schema is concatenated in text for running the inference process, while the computed Matched Schema cache is then stored in the schema cache pool.
\end{itemize}

Overall, by decomposing the information extraction task into two stages—Identification and Extraction—the multi-LoRA structure of on-device LLMs can be fully leveraged to optimize each stage separately, thereby improving overall extraction performance.
By incrementally caching previously inferenced schemas, we can avoid redundant calculations, thereby boosting the extraction inference efficiency.

\section{Experiments}

\subsection{Experimental Settings}

\paragraph{RAG-based Baselines.}
We compare our proposed DLISC approach with the following retrieval-augmented baselines (as shown in Figure~\ref{fig:architecture} (a), K=5):
\begin{itemize}[leftmargin=*, itemsep=-2pt, topsep=2pt, parsep=2pt]
\item \textbf{BM25} \cite{DBLP:journals/ftir/RobertsonZ09} is a probabilistic information retrieval algorithm that scores document-query relevance by weighting term frequency, inverse document frequency, and document length.
\item \textbf{BGE-Reranker-V2-M3}~\cite{bgem3} is a lightweight reranker model that possesses strong multilingual capabilities, is easy to deploy, and supports fast inference.
\item \textbf{LLM-Embedder} \cite{zhang-etal-2024-multi-task} comprehensively support diverse retrieval augmentation scenarios for LLMs with a unified embedding model, addressing the limitations of both general-purpose and task-specific retrievers.
\end{itemize}


\paragraph{On-Device LLMs.}

We consider the following state-of-the-art on-device LLMs for information extraction:

\begin{itemize}[leftmargin=*, itemsep=-2pt, topsep=2pt, parsep=2pt]
    \item \textbf{Llama-3.2-1B}~\cite{Dubey2024TheL3} delivers powerful language model capabilities on edge and mobile devices with its lightweight 1B parameter model.
    \item \textbf{Qwen2.5-3B} \cite{Yang2024Qwen2TR} demonstrate exceptional performance across a wide range of tasks and benchmarks, showcasing its strength in instruction following, generating long texts, understanding structured data, and producing structured outputs.
    \item \textbf{TinyLlama-1.1B-Chat-v1.0} \cite{Zhang2024TinyLlamaAO} is a lightweight conversational model based on the TinyLlama project, which aims to pretrain a 1.1 billion parameter Llama model on 3 trillion tokens and is suitable for applications with limited computational and memory resources. 
\end{itemize}

\paragraph{Datasets.} We conduct experiments on several collected datasets, including: 
\begin{itemize}[leftmargin=*, itemsep=-2pt, topsep=0pt, parsep=2pt]
    \item CrossNER\_AI \cite{Liu2020CrossNEREC}: CrossNER is a well-known \textbf{English} open-source project in the field of NLP, specifically focusing on cross-domain Named Entity Recognition (NER). In particular, CrossNER\_AI mainly focuses on the \textbf{artificial intelligence (AI)} domain.
    \item  DuEE-Fin \cite{gui-etal-2024-iepile}: a large-scale dataset designed for document-level Event Extraction (EE) tasks, particularly focusing on the \textbf{Chinese financial} domain.
\end{itemize}

\paragraph{Evaluation Metrics.} We use three metrics to evaluate IE performance in terms of both effectiveness and efficiency:
\begin{itemize}[leftmargin=*, itemsep=-2pt, topsep=0pt, parsep=2pt]
    \item Precision – Assesses the accuracy of schema matching.
    \item F1 Score – Measures the overall quality of information extraction.
    \item Latency (seconds, s) – Captures efficiency by recording average extraction time over 100 samples.
\end{itemize}

\begin{table}[!t]
    \centering
    \addtolength{\tabcolsep}{-0pt}
    \resizebox{0.44\textwidth}{!}{
    \begin{tabular}{lccc}
        \toprule
        \textbf{Method} & \textbf{Llama-3.2-1B} & \textbf{Qwen2.5-3B} \\
        \midrule
        BM25 & 0.2341 & 0.3968 \\
        BGE-Reranker-V2-M3 & 0.2341 & 0.3819 \\
        LLM-Embedder & 0.2341 & 0.3819 \\
        \model & \textbf{0.4179} & \textbf{0.4311} \\
        \bottomrule
    \end{tabular}}
    \caption{Effectiveness of DLISC with baselines on CrossNER\_AI for NER in terms of F1.}
    \label{tab:ner_llm_comparison}
\end{table}

\begin{table}[!t]
    \centering
    \addtolength{\tabcolsep}{-0pt}
    \resizebox{0.36\textwidth}{!}{
    \begin{tabular}{lcc}
        \toprule
        \textbf{Method} & \textbf{Llama-3.2-1B} & \textbf{Qwen2.5-3B} \\
        \midrule
        Dual-LoRA & 1.38 & 6.02 \\
        \model & \textbf{0.66} & \textbf{4.59} \\
        \bottomrule
    \end{tabular}}
    \caption{Efficiency of Dual-LoRA without caching and \model\ in seconds (s) per sample.}
    \label{tab:efficiency}
\end{table}

\begin{table}[!t]
\centering
\addtolength{\tabcolsep}{-0pt}
\resizebox{0.44\textwidth}{!}{
\begin{tabular}{lcc}
\toprule
\textbf{Method} & \textbf{Identification} & \textbf{Extraction} \\
& \textbf{(Precision)}& \textbf{(F1)} \\
\midrule
BM25 & 0.2120 & 0.3220 \\
BGE-Reranker-V2-M3 & 0.2580 & 0.5714 \\
LLM-Embedder & 0.2680 & 0.7484 \\
\model & \textbf{0.2982} & \textbf{0.7746} \\
\bottomrule
\end{tabular}}
\caption{Ablation exploration on DuEE-Fin dataset with TinyLlama-1.1B-Chat-v1.0 as the base LLM $(\boldsymbol{\theta})$. ``Identification'' represents the schema matching phase, and ``Extraction'' represents the end-to-end information extraction phase with the corresponding matched schemas.}
\label{tab:ir_metrics}
\vspace{-1\baselineskip}
\end{table}

\subsection{Experimental Results}

\paragraph{Effectiveness Comparison.}

Table~\ref{tab:ner_llm_comparison} presents a comparative evaluation between our proposed DLISC method and several retrieval-augmented information extraction (IE) baselines that differ in their retrieval capabilities. Specifically, the baselines incorporate three retrieval models—BM25, BGE-Reranker-V2-M3, and LLM-Embedder—while employing the same information extraction backbone model $\boldsymbol{\theta}_E^\prime$ as used in DLISC, ensuring a fair comparison. As shown in the results, DLISC consistently outperforms all RAG-based baselines across both Llama-3.2-1B and Qwen2.5-3B, demonstrating superior extraction effectiveness. These findings highlight that DLISC’s on-device IE task decomposition and schema caching strategies contribute to more accurate and robust information extraction, especially in complex or schema-rich scenarios.

\paragraph{Efficiency Comparison.}

To assess the impact of Incremental Schema Caching on extraction efficiency, we measure the end-to-end processing time (in seconds) required for information extraction over a set of 100 test samples. 
As shown in Table~\ref{tab:efficiency}, we compare the baseline Dual-LoRA method without caching against our proposed DLISC approach enhanced with Incremental Schema Caching. 
Notably, DLISC, when implemented with both Llama-3.2-1B and Qwen2.5-3B, achieves a substantial reduction in latency, demonstrating a more efficient inference process.
The results highlight that Incremental Schema Caching notably lowers the average extraction time per sample, thereby improving overall system responsiveness.

\paragraph{Ablation Exploration.}

Our proposed DLISC adopts a two-stage information extraction paradigm involving two distinct LLMs: $\boldsymbol{\theta}_I^\prime$ for identifying the most relevant schemas, and $\boldsymbol{\theta}_E^\prime$ for schema-aware extraction.
Table~\ref{tab:ir_metrics} analyzes the contribution of each component—namely the ``Identification'' and ``Extraction'' phases—on the DuEE-Fin dataset, using TinyLlama-1.1B-Chat-v1.0 as the base LLM ($\boldsymbol{\theta}$).
The results show that DLISC outperforms all three RAG-based IE baselines, achieving the highest Precision score in the Identification phase and the highest F1 score in the Extraction phase.
Overall, DLISC delivers the best end-to-end performance, validating the effectiveness of the Dual-LoRA information extraction paradigm.

\section{Conclusion}

In this paper, we propose \underline{D}ual-\underline{L}oRA with \underline{I}ncremental \underline{S}chema \underline{C}aching (DLISC), a novel two-stage information extraction approach tailored for on-device LLMs.
Specifically, DLISC employs an Identification LoRA module to retrieve the most relevant schemas for a given query, and an Extraction LoRA module to perform information extraction conditioned on the selected schemas.
We conduct extensive experiments on multiple benchmark datasets, demonstrating that DLISC achieves state-of-the-art performance in both schema identification and schema-aware extraction when compared with three RAG-based IE baselines.
To further improve inference efficiency, DLISC integrates Incremental Schema Caching, which effectively reduces redundant computation.
Future work will explore integrating more fine-grained schema representations and dynamic on-demand generation to further enhance adaptability across diverse extraction tasks. 

\section{Limitations}

Our Dual-LoRA with Incremental Schema Caching (DLISC) framework has several limitations.
First, DLISC currently employs only the LoRA adapters, so it remains to be explored whether the approach can generalize to other adapter types~\cite{hu-etal-2023-llm}, such as Prefix Tuning, Series Adapter, or Parallel Adapter, to improve adaptability and generalization.
Second, we were unable to deploy on real edge devices due to computational resource constraints.
Additionally, our current implementation lacks support for more complex multilingual and cross-lingual scenarios, posing additional challenges for building scalable and versatile information extraction systems.
We hope to address the above limitations in the follow-up work.


\bibliography{paper}

\begin{thebibliography}{20}
\providecommand{\natexlab}[1]{#1}

\bibitem[{Chen et~al.(2024)Chen, Xiao, Zhang, Luo, Lian, and Liu}]{bgem3}
Jianlv Chen, Shitao Xiao, Peitian Zhang, Kun Luo, Defu Lian, and Zheng Liu. 2024.
\newblock \href {https://arxiv.org/abs/2402.03216} {Bge m3-embedding: Multi-lingual, multi-functionality, multi-granularity text embeddings through self-knowledge distillation}.
\newblock \emph{Preprint}, arXiv:2402.03216.

\bibitem[{Deng et~al.(2024)Deng, Ma, Zhang, Cao, and Hooi}]{deng2024information}
Shumin Deng, Yubo Ma, Ningyu Zhang, Yixin Cao, and Bryan Hooi. 2024.
\newblock Information extraction in low-resource scenarios: Survey and perspective.
\newblock In \emph{2024 IEEE International Conference on Knowledge Graph (ICKG)}, pages 33--49. IEEE.

\bibitem[{Dubey et~al.(2024)Dubey, Jauhri, and Pandey}]{Dubey2024TheL3}
Abhimanyu Dubey, Abhinav Jauhri, and Abhinav Pandey. 2024.
\newblock \href {https://api.semanticscholar.org/CorpusID:271571434} {The llama 3 herd of models}.
\newblock \emph{ArXiv}, abs/2407.21783.

\bibitem[{Gao et~al.(2023)Gao, Xiong, Gao, Jia, Pan, Bi, Dai, Sun, and Wang}]{gao2023retrieval}
Yunfan Gao, Yun Xiong, Xinyu Gao, Kangxiang Jia, Jinliu Pan, Yuxi Bi, Yi~Dai, Jiawei Sun, and Haofen Wang. 2023.
\newblock Retrieval-augmented generation for large language models: A survey.
\newblock \emph{arXiv preprint arXiv:2312.10997}.

\bibitem[{Gim et~al.(2024)Gim, Chen, Lee, Sarda, Khandelwal, and Zhong}]{gim2024prompt}
In~Gim, Guojun Chen, Seung-seob Lee, Nikhil Sarda, Anurag Khandelwal, and Lin Zhong. 2024.
\newblock Prompt cache: Modular attention reuse for low-latency inference.
\newblock \emph{Proceedings of Machine Learning and Systems}, 6:325--338.

\bibitem[{Gui et~al.(2024)Gui, Yuan, Ye, Zhang, Sun, Liang, and Chen}]{gui-etal-2024-iepile}
Honghao Gui, Lin Yuan, Hongbin Ye, Ningyu Zhang, Mengshu Sun, Lei Liang, and Huajun Chen. 2024.
\newblock \href {2024.acl-short.13} {{IEP}ile: Unearthing large scale schema-conditioned information extraction corpus}.
\newblock pages 127--146, Bangkok, Thailand.

\bibitem[{Hu et~al.(2023)Hu, Wang, Lan, Xu, Lim, Bing, Xu, Poria, and Lee}]{hu-etal-2023-llm}
Zhiqiang Hu, Lei Wang, Yihuai Lan, Wanyu Xu, Ee-Peng Lim, Lidong Bing, Xing Xu, Soujanya Poria, and Roy Lee. 2023.
\newblock \href {https://doi.org/10.18653/v1/2023.emnlp-main.319} {{LLM}-adapters: An adapter family for parameter-efficient fine-tuning of large language models}.
\newblock In \emph{Proceedings of the 2023 Conference on Empirical Methods in Natural Language Processing}, pages 5254--5276, Singapore. Association for Computational Linguistics.

\bibitem[{Li et~al.(2024)Li, Jin, Zhou, Zhang, Zhang, Zhu, and Dou}]{Li2024FromMT}
Xiaoxi Li, Jiajie Jin, Yujia Zhou, Yuyao Zhang, Peitian Zhang, Yutao Zhu, and Zhicheng Dou. 2024.
\newblock \href {https://api.semanticscholar.org/CorpusID:269303210} {From matching to generation: A survey on generative information retrieval}.
\newblock \emph{ArXiv}, abs/2404.14851.

\bibitem[{Liang et~al.(2025)Liang, Lv, Wen, Wu, Zhang, Wang, and Liu}]{liang2025adaptive}
Sheng Liang, Hang Lv, Zhihao Wen, Yaxiong Wu, Yongyue Zhang, Hao Wang, and Yong Liu. 2025.
\newblock Adaptive schema-aware event extraction with retrieval-augmented generation.
\newblock \emph{arXiv preprint arXiv:2505.08690}.

\bibitem[{Liu et~al.(2020)Liu, Xu, Yu, Dai, Ji, Cahyawijaya, Madotto, and Fung}]{Liu2020CrossNEREC}
Zihan Liu, Yan Xu, Tiezheng Yu, Wenliang Dai, Ziwei Ji, Samuel Cahyawijaya, Andrea Madotto, and Pascale Fung. 2020.
\newblock \href {https://api.semanticscholar.org/CorpusID:227736891} {Crossner: Evaluating cross-domain named entity recognition}.
\newblock In \emph{AAAI Conference on Artificial Intelligence}.

\bibitem[{Luohe et~al.(2024)Luohe, Zhang, Yao, Li, and Hai}]{Luohe2024KeepTC}
Shi Luohe, Hongyi Zhang, Yao Yao, Z.~Li, and Zhao Hai. 2024.
\newblock \href {https://api.semanticscholar.org/CorpusID:271432418} {Keep the cost down: A review on methods to optimize llm' s kv-cache consumption}.
\newblock \emph{ArXiv}, abs/2407.18003.

\bibitem[{Mehta et~al.(2024)Mehta, Sekhavat, Cao, Horton, Jin, Sun, Mirzadeh, Najibi, Belenko, Zatloukal et~al.}]{mehta2024openelm}
Sachin Mehta, Mohammad~Hossein Sekhavat, Qingqing Cao, Maxwell Horton, Yanzi Jin, Chenfan Sun, Iman Mirzadeh, Mahyar Najibi, Dmitry Belenko, Peter Zatloukal, et~al. 2024.
\newblock Openelm: An efficient language model family with open training and inference framework.
\newblock \emph{arXiv preprint arXiv:2404.14619}.

\bibitem[{Robertson and Zaragoza(2009)}]{DBLP:journals/ftir/RobertsonZ09}
Stephen~E. Robertson and Hugo Zaragoza. 2009.
\newblock \href {https://doi.org/10.1561/1500000019} {The probabilistic relevance framework: {BM25} and beyond}.
\newblock \emph{Found. Trends Inf. Retr.}, 3(4):333--389.

\bibitem[{Shiri et~al.(2024)Shiri, Nguyen, Moghimifar, Yoo, Haffari, and Li}]{shiri-2024-decompose}
Fatemeh Shiri, Van Nguyen, Farhad Moghimifar, John Yoo, Gholamreza Haffari, and Yuan-Fang Li. 2024.
\newblock \href {https://arxiv.org/abs/2406.01045} {Decompose, enrich, and extract! schema-aware event extraction using llms}.
\newblock \emph{Preprint}, arXiv:2406.01045.

\bibitem[{Xu et~al.(2023)Xu, Chen, Peng, Zhang, Xu, Zhao, Wu, Zheng, and Chen}]{xu2023large}
Derong Xu, Wei Chen, Wenjun Peng, Chao Zhang, Tong Xu, Xiangyu Zhao, Xian Wu, Yefeng Zheng, and Enhong Chen. 2023.
\newblock Large language models for generative information extraction: A survey.
\newblock \emph{arXiv preprint arXiv:2312.17617}.

\bibitem[{Xu et~al.(2024)Xu, Li, Chen, Wang, Gao, Cai, and Ling}]{xu2024device}
Jiajun Xu, Zhiyuan Li, Wei Chen, Qun Wang, Xin Gao, Qi~Cai, and Ziyuan Ling. 2024.
\newblock On-device language models: A comprehensive review.
\newblock \emph{arXiv preprint arXiv:2409.00088}.

\bibitem[{Yang et~al.(2024)Yang, Yang, and Hui}]{Yang2024Qwen2TR}
An~Yang, Baosong Yang, and Binyuan Hui. 2024.
\newblock \href {https://api.semanticscholar.org/CorpusID:271212307} {Qwen2 technical report}.
\newblock \emph{ArXiv}, abs/2407.10671.

\bibitem[{Yang et~al.(2022)Yang, Wu, Yang, Lian, Guo, and Wang}]{yang2022survey}
Yang Yang, Zhilei Wu, Yuexiang Yang, Shuangshuang Lian, Fengjie Guo, and Zhiwei Wang. 2022.
\newblock A survey of information extraction based on deep learning.
\newblock \emph{Applied Sciences}, 12(19):9691.

\bibitem[{Zhang et~al.(2024{\natexlab{a}})Zhang, Liu, Xiao, Dou, and Nie}]{zhang-etal-2024-multi-task}
Peitian Zhang, Zheng Liu, Shitao Xiao, Zhicheng Dou, and Jian-Yun Nie. 2024{\natexlab{a}}.
\newblock \href {https://doi.org/10.18653/v1/2024.acl-long.194} {A multi-task embedder for retrieval augmented {LLM}s}.
\newblock In \emph{Proceedings of the 62nd Annual Meeting of the Association for Computational Linguistics (Volume 1: Long Papers)}, pages 3537--3553, Bangkok, Thailand. Association for Computational Linguistics.

\bibitem[{Zhang et~al.(2024{\natexlab{b}})Zhang, Zeng, Wang, and Lu}]{Zhang2024TinyLlamaAO}
Peiyuan Zhang, Guangtao Zeng, Tianduo Wang, and Wei Lu. 2024{\natexlab{b}}.
\newblock \href {https://api.semanticscholar.org/CorpusID:266755802} {Tinyllama: An open-source small language model}.
\newblock \emph{ArXiv}, abs/2401.02385.

\end{thebibliography}



\end{document}